\documentclass[10pt,twocolumn,letterpaper]{article}

\usepackage{iccv}
\usepackage{times}
\usepackage{epsfig}
\usepackage{graphicx}
\usepackage{amsmath}
\usepackage{amssymb}
\usepackage{url} 
\usepackage{bigstrut}  
\usepackage{multirow}
\usepackage{comment}
\usepackage{subcaption}
\usepackage{blindtext}
\usepackage{enumitem}
\usepackage{booktabs}
\usepackage{epstopdf}
\usepackage[normalem]{ulem}
\usepackage{color}
\usepackage{tabularx}
\usepackage{lipsum}

\usepackage[pagebackref=true,breaklinks=true,letterpaper=true,colorlinks,bookmarks=false]{hyperref}

\iccvfinalcopy 

\ificcvfinal\fi
\begin{document}
\pagenumbering{gobble} 
\title{Human Pose Estimation using Global and Local Normalization}

\author{{Ke Sun{\small $~^{1}$}}, ~Cuiling Lan{\small $~^{2}$}, ~Junliang Xing{\small $~^{3}$}, Wenjun Zeng{\small $~^{2}$}, ~Dong Liu{\small $~^{1}$}, Jingdong Wang{\small $~^{2}$}\\
	\normalsize
	$^{1}$\	University of Science and Technology of China, Anhui, China ~~ $^{2}$\,Microsoft Research Asia, Beijing, China\\
	\normalsize
	$^{3}$\,National Laboratory of Pattern Recognition, Institute of Automation, Chinese Academy of Sciences, Beijing, China \\
	\normalsize
	sunk@mail.ustc.edu.cn,
	\{culan,wezeng,jingdw\}@microsoft.com, jlxing@nlpr.ia.ac.cn,  
	dongeliu@ustc.edu.cn
}

\maketitle
\newcommand\blfootnote[1]{%
	\begingroup 
	\renewcommand\thefootnote{}\footnote{#1}%
	\addtocounter{footnote}{-1}%
	\endgroup 
}
\blfootnote{This work was done when Ke Sun was an intern at Microsoft Research Asia. Junliang Xing is partly supported by the Natural Science Foundation of China (Grant No. 61672519).}

\begin{abstract}
	In this paper, we address the problem of estimating the positions of human joints, \textit{i.e.}, articulated pose estimation. Recent state-of-the-art solutions model two key issues, joint detection and spatial configuration refinement, together using convolutional neural networks. Our work mainly focuses on spatial configuration refinement by reducing variations of human poses statistically, which is motivated by the observation that the scattered distribution of the relative locations of joints (\textit{e.g.}, the left wrist is distributed nearly uniformly in a circular area around the left shoulder) makes the learning of convolutional spatial models hard. We present a two-stage normalization scheme, human body normalization and limb normalization, to make the distribution of the relative joint locations compact, resulting in easier learning of convolutional spatial models and more accurate pose estimation. In addition, our empirical results show that incorporating multi-scale supervision and multi-scale fusion into the joint detection network is beneficial. Experiment results demonstrate that our method consistently outperforms state-of-the-art methods on the benchmarks.
\end{abstract}

\begin{figure}[ht]
	\begin{subfigure}{0.5\textwidth}
		\centering
		\includegraphics[width=0.95\linewidth]{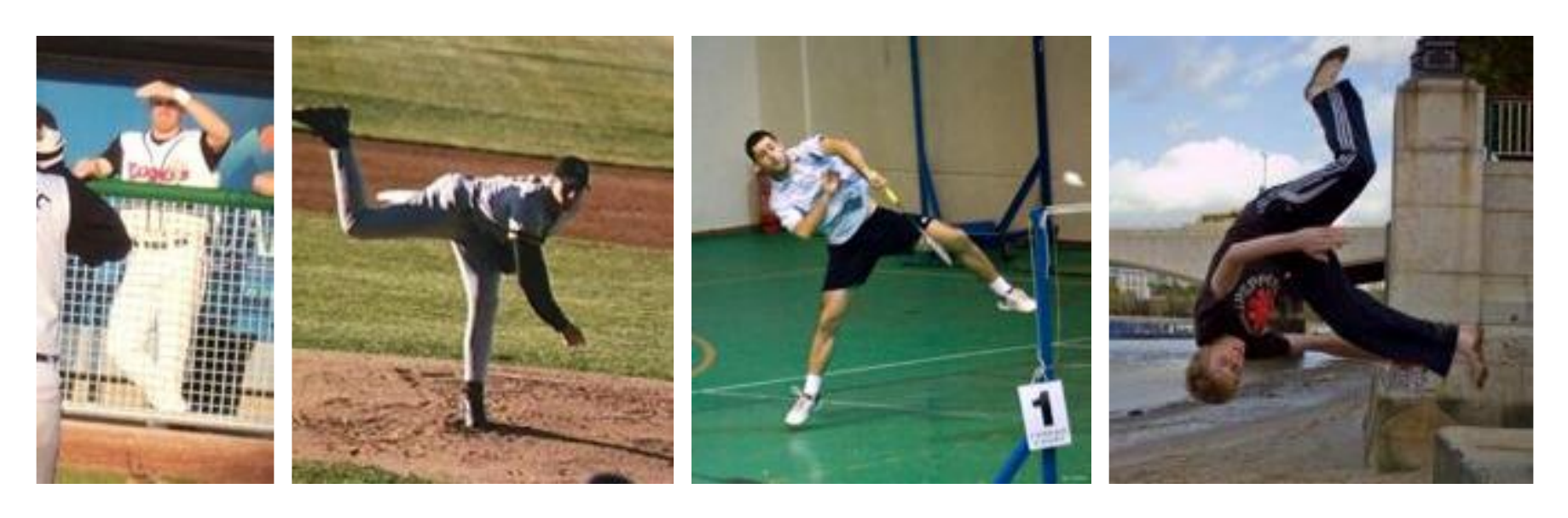}
		\vspace{-2mm}
		\caption{}
		\label{fig:sfiga}
	\end{subfigure}
	\begin{subfigure}{0.236\textwidth}
		\vspace{-1mm}
		\centering
		\includegraphics[width=1\linewidth]{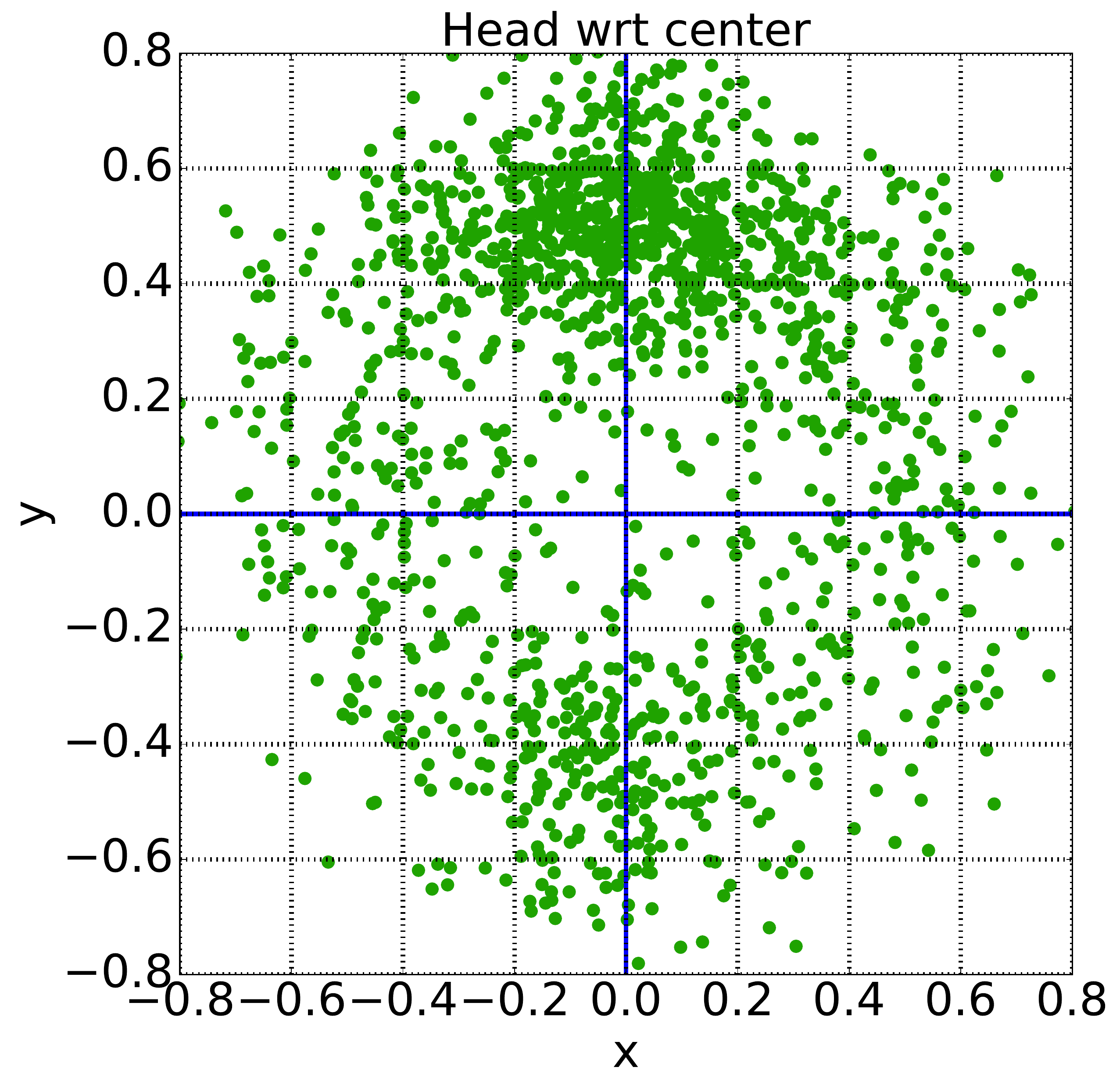} 
		\vspace{-5mm}
		\caption{}
		\label{fig:sfigb}
	\end{subfigure}	
	\begin{subfigure}{0.236\textwidth}
		\vspace{-1mm}
		\centering
		\includegraphics[width=1\linewidth]{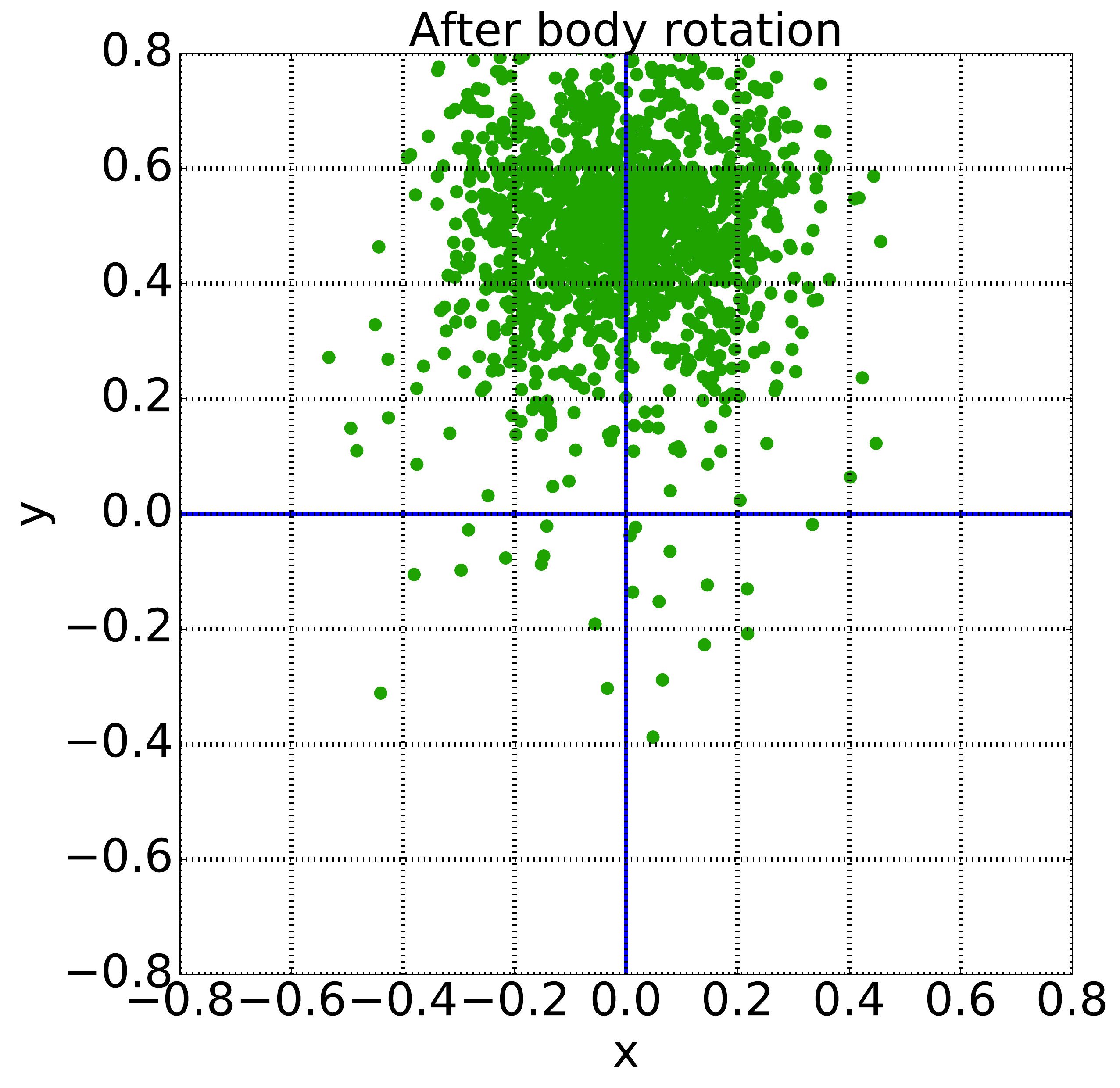} 
		\vspace{-5mm}
		\caption{}
		\label{fig:sfigc}
	\end{subfigure}
	\begin{subfigure} {0.236\textwidth}
		\vspace{-1mm}
		\centering
		\includegraphics[width=1\linewidth]{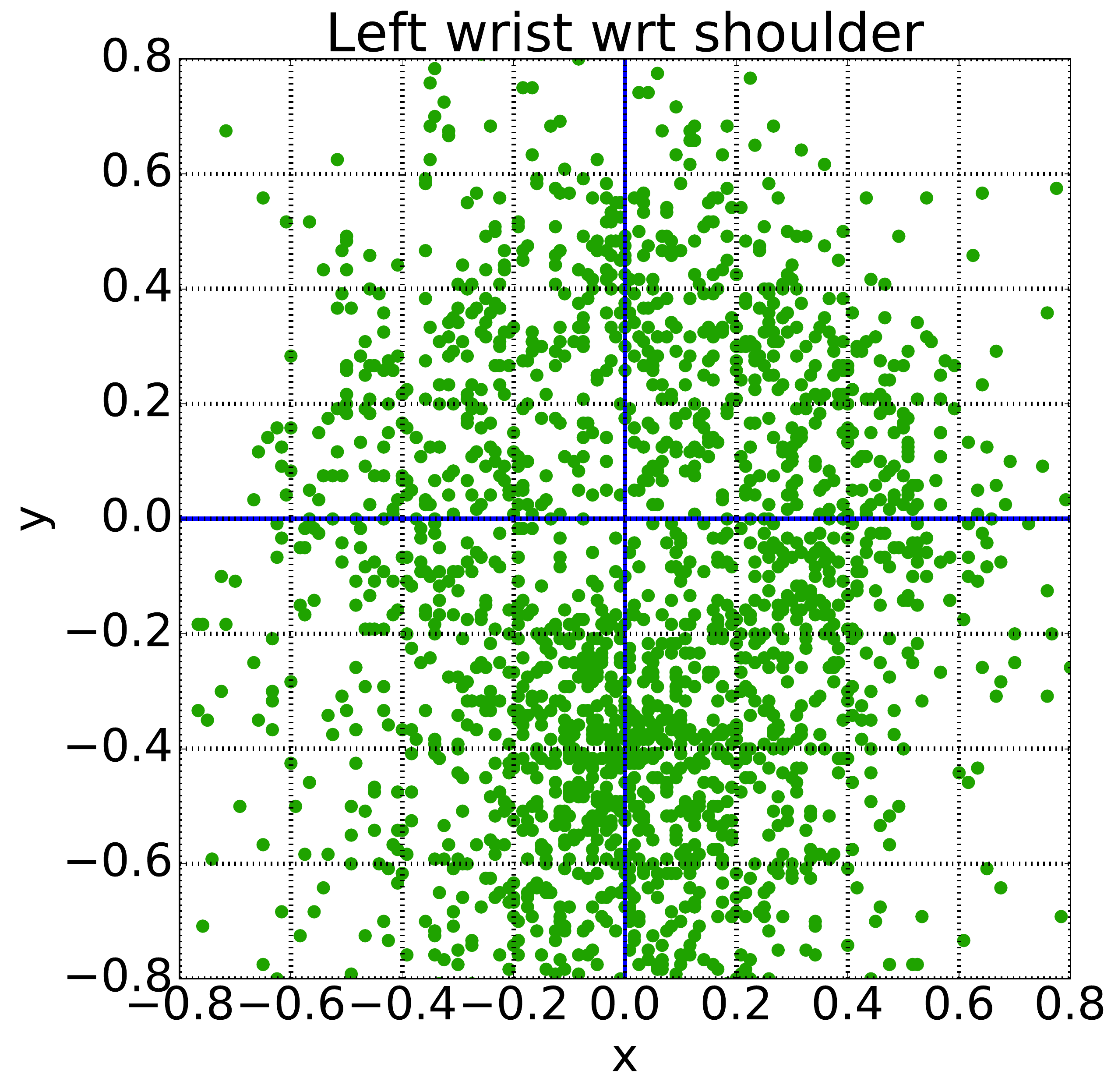}
		\vspace{-5mm}
		\caption{}
		\label{fig:sfigd}
	\end{subfigure}
	\begin{subfigure}{0.236\textwidth}
		\centering
		\includegraphics[width=1\linewidth]{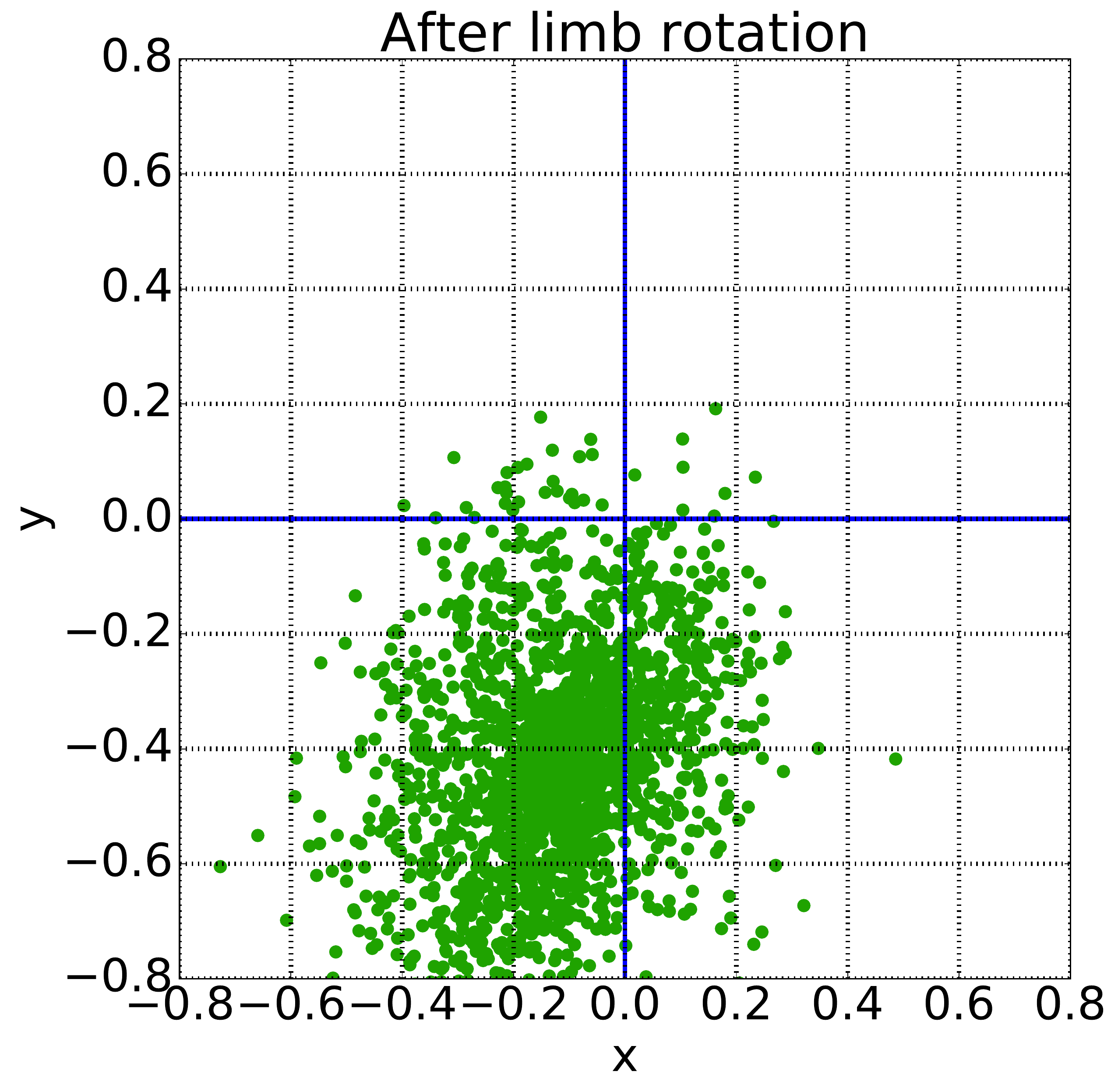}
		\vspace{-5mm}
		\caption{}
		\label{fig:sfige}
	\end{subfigure}	
	\vspace{-2mm}
	\caption{Pose normalization can compact the relative position distribution. (a) Example images with various poses from LSPET~\cite{Johnson2011LSP}. (b) and (d) are originally relative positions of two joints. (b) shows the positions of heads with respect to the body centers. (d) shows the positions of the left wrists with respect to the left shoulders. (c) and (e) are the relative positions after body and limb normalization corresponding to (b) and (d) respectively. The distributions of the relative positions in (c) and (e) are much more compact. 
	} \label{fig:diversity}
\end{figure}

\section{Introduction}
Human pose estimation is one of the most challenging problems in computer vision and plays an essential role in human body modeling. It has wide applications such as human action recognition \cite{xiaohan2015joint}, activity analyses \cite{aggarwal2011human}, and human-computer interaction \cite{shotton2013real}. Despite many years of research with significant progress made recently \cite{andriluka2009pictorial, eichner2009better, dantone2013human, Chen_NIPS14, tompson2014joint, tompson2015efficient}, pose estimation still remains a very challenging task, mainly due to the large variations in body postures, shapes, complex inter-dependency of parts, clothing and so on.

There are two key problems in pose estimation: joint detection, which estimates the confidence level of each pixel being some joint, and spatial refinement, which usually refines the confidence for joints by exploiting the spatial configuration of the human body.

Our work follows this path and mainly focuses on spatial configuration refinement. It is observed that the distribution of relative positions of joints may be very diverse with respect to their neighboring ones. Examples regarding the distributions of the joints on the LSPET dataset \cite{Johnson2011LSP} are shown in Figure~\ref{fig:diversity}. The relative positions of the head with respect to the center of the human body are shown in Figure~\ref{fig:diversity} (b), which is distributed almost uniformly in a circular region. After making the human body upright, the distribution becomes much more compact, as shown in Figure~\ref{fig:diversity} (c). We have similar observations for other neighboring joints. For some joints on the limbs (\eg, wrist, ankle), their distributions are still diverse even after positioning the torso upright. We further rotate the human upper limb (\eg, the left arm) to a vertical downward positions. The distribution of the relative positions of the left wrist, shown in Figure~\ref{fig:diversity} (e), becomes much more compact.

The diversity of orientations (\eg, body and limb) is the main factor in the variations of pose. Motivated by these observations, we propose two normalization schemes, reducing diversity to generate compact distributions. The first normalization scheme is human body normalization, rotating the human body to upright according to joint detection results, which globally makes the relative positions between joints compactly distributed. This scheme is followed by a global spatial refinement module to refine all the estimations of the joints. The second one is limb normalization: rotating the joints of each limb to make the relative positions more compact. There are four total limb normalization modules, and each is followed by a spatial limb refinement module to refine the estimations from the global spatial refinement. Thanks to the normalization schemes, a much more consistent spatial configuration of the human body can be obtained, which facilitates the learning of spatial refinement models.

Besides the observations in ~\cite{Lee2015, long2015fully, szegedy2015googlenet,  xie2015holistically} that the multi-stage supervision, \eg, supervision on the joint detection stage and the spatial refinement stage, is helpful, we observe that multi-scale supervision and multi-scale fusion over the convolutional network within the joint detection stage are also beneficial.


Our main contribution lies in effective normalization schemes to facilitate the learning of convolutional spatial models. Our scheme can be applied following different joint detectors for refining the spatial configurations. Our experiment results demonstrate the effectiveness on several joint detectors, such as FCN~\cite{long2015fully}, ResNet~\cite{he2016Resi} and Hourglass~\cite{hourglass16}. An additional minor contribution is that we empirically show the improvement by using an architecture with multi-scale supervision and fusion for joint detection.

\section{Related Work}
Significant progress has been made recently in human pose estimation by deep learning based methods \cite{toshev2014deeppose, ouyang2014multi, Chen_NIPS14, tompson2015efficient, pishchulin16cvpr, insafutdinov16eccv, hourglass16}. The joint detection and joint relation models are widely recognized as two key components in solving this problem. In the following, we briefly review related developments on these two components respectively and discuss some related works which motivate our design of the normalization scheme.

\textbf{Joint detection model.} Many recent works use convolutional neural networks to learn feature representations for obtaining the score maps of joints or the locations of joints \cite{toshev2014deeppose,fan2015combing,IEF16,wei2016cpm,hourglass16,BMVC16pose,regression2016pose}. Some methods directly employ learned feature representations to regress joint positions, \eg, the DeepPose method \cite{toshev2014deeppose}. A more typical way of joint detection is to estimate a score map for each joint based on the fully convolutional neural network (FCN) \cite{long2015fully}. The estimation procedure can be formulated as a multi-class classification problem \cite{wei2016cpm,hourglass16} or regression problem \cite{IEF16,tompson2015efficient}. For the multi-class formulation, either a single-label based loss (\eg, softmax cross-entropy loss) \cite{chu2016structure} or a multi-label based loss (\eg, sigmoid cross-entropy loss) \cite{pishchulin16cvpr} can be used. One main problem for the FCN-based joint detection model is that the positions of joints are estimated from low resolution score maps. This reduces the location accuracy of the joints. In our work, we introduce multi-scale supervision and fusion to further improve performance with gradual up-sampling.

\textbf{Joint relation model.} The pictorial structures \cite{yang2011articulated, pishchulin2013poselet} define the deformable configurations by spring-like connections between pairs of parts to model complex joint relations. Subsequent works \cite{ouyang2014multi, Chen_NIPS14,yang2016end} extend such an idea to convolutional neuron networks. In those approaches, to model the human poses with large variations, a mixture model is usually learned for each joint. Tompson et al. \cite{tompson2014joint} formulates the spatial relations as a Markov Random Field (MRF) like model over the distribution of spatial locations for each body part. The location distribution of one joint relative to another is modeled by convolutional prior which is expected to give some spatial predictions and remove false positive outliers for each joint. Similarly, the structured feature learning method in \cite{chu2016structure} adapts geometrical transform kernels to capture the spatial relationships of joints from feature maps. To better estimate the human pose, complex network architecture design with many more parameters are expected on account of the articulated structure of the human body, such as \cite{chu2016structure} and \cite{yang2016end}. In our work, we address this problem by compensating for the variations of poses both globally and locally for facilitating the spatial configuration exploration.

\begin{figure*}[t]
	\centering
	\includegraphics[width=1\linewidth]{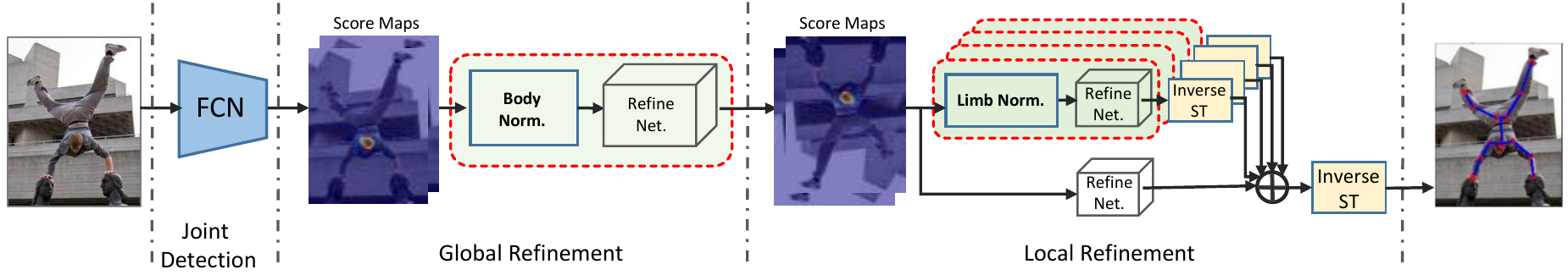}
	\caption{Proposed framework with global and local normalization. Joint detection with fully convolutional network (FCN) provides initial estimation of joints in terms of score maps. In the global refinement stage, a body (global) normalization module rotates the score maps to have upright position for the body, followed by a refinement module. In the local refinement stage, limb (local) normalization modules rotates the score maps to have vertical downward position for limbs, followed by refinements.}
	\label{fig:framework}
\end{figure*}

\textbf{Normalization.} Normalization of the training samples to reduce their variations has been proven to be a key step before learning models using these samples. For example, in the PCA Whitening and ZCA Whitening operations \cite{kessy2015optimal}, the feature pre-processing step are adapted before training an SVM classifier, etc. The batch normalization \cite{icml2015_batchnorm} technique accelerates the deep network training by reducing the internal covariate shift across layers. In computer vision applications, face normalization has been found to be very helpful for improving the face recognition performance \cite{asthana2011fully,zhu2015high, chen2016face}. It is beneficial for decreasing the intra-person variations and achieving pose-invariant recognition. 

\section{Our Approach}
\label{ss_approach}
Human pose estimation is defined as the problem of localization of human joints, \ie, head, neck \etal, from an image. Given an image $I$, the goal is to estimate the positions of the joints: $\{(x_k,y_k)\}_{k=1}^{K}$, where $K$ is the number of joints.

\subsection{Pipeline}
Figure~\ref{fig:framework} shows the framework of our proposed approach. It consists of joint detection and spatial configuration refinement, which are both realized with convolutional neural networks. The output of joint detector consists of $K+1$ score maps, including $K$ joint score maps, providing spatial configuration information and one non-joint (background) score map. The value in each score map indicates the degree of confidence that the pixel is the corresponding joint. With the score maps generated by the former stage (\eg, joint detector or refinement stage) as the input, two normalization stages correct wrongly predicted joints based on spatial configurations of the human body. Note that we focus on the exploration of the spatial configurations of joints for refinement. Unlike many other works \cite{regression2016pose, hourglass16, wei2016cpm}, we do not incorporate low level features and our refinement is based on the score maps which indicate the probabilities of being each joint. 

\subsection{Spatial Configuration Refinement}
There are two stages for spatial configuration refinement as depicted in Figure~\ref{fig:framework}. The first stage is a global refinement, consisting of a global normalization module and a refinement module that refines all $K$ joints. The second stage includes two parallel refinement modules: semi-global refinement and local refinement. The local refinement module consists of four branches. Each branch corresponds to a limb and contains a local limb normalization module and a local refinement module. Inverse normalizations by inverse spatial transforms are used to rotate the joints/body back for obtaining the final results. 

\begin{figure}[ht]
	\begin{subfigure}{0.11\textwidth} 
		\centering
		\includegraphics[width=0.72\linewidth]{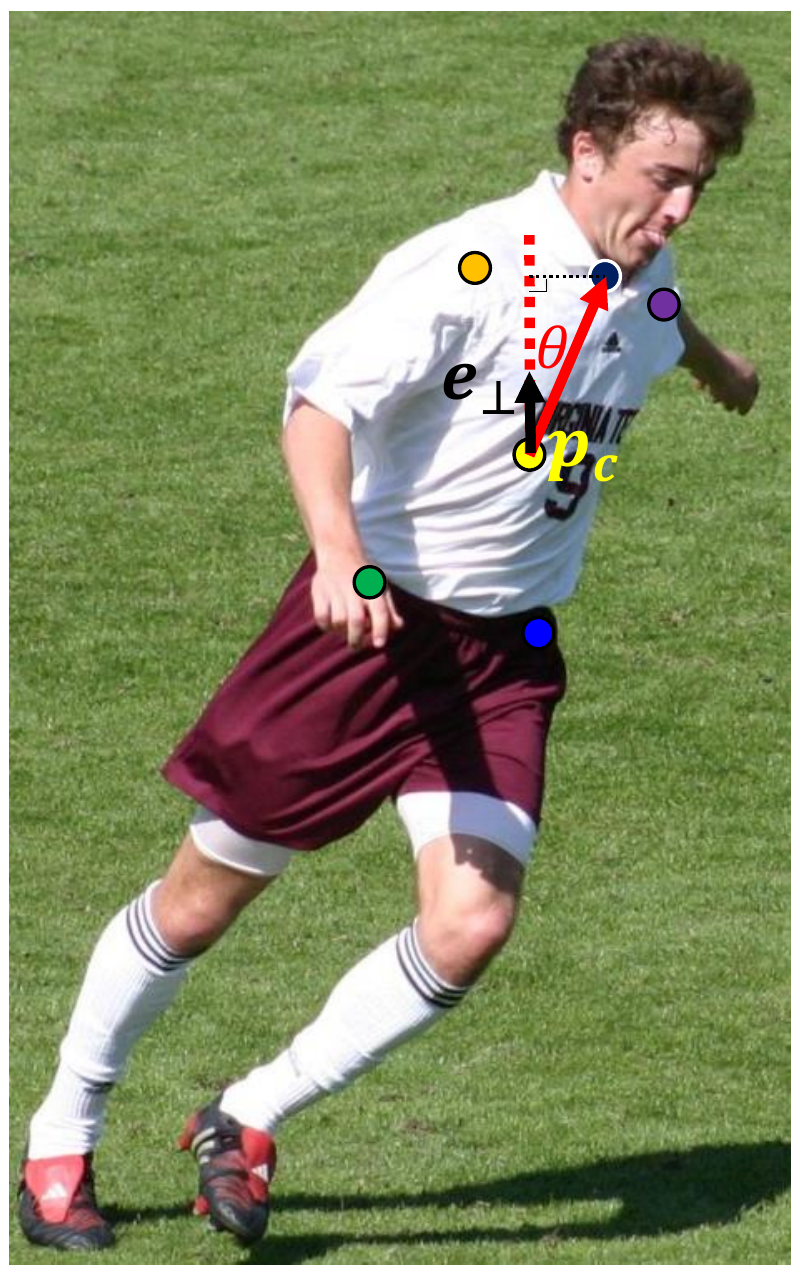}
		\vspace{-1mm}
		\caption{}
		\label{fig:s3figa}
	\end{subfigure}
	\begin{subfigure}{0.11\textwidth}
		\centering
		\includegraphics[width=0.92\linewidth]{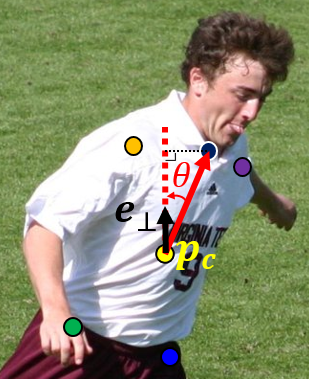} 
		\vspace{-1mm}
		\caption{}
		\label{fig:s3figb}
	\end{subfigure}
	\begin{subfigure}{0.11\textwidth} 
		\centering
		\includegraphics[width=0.86\linewidth]{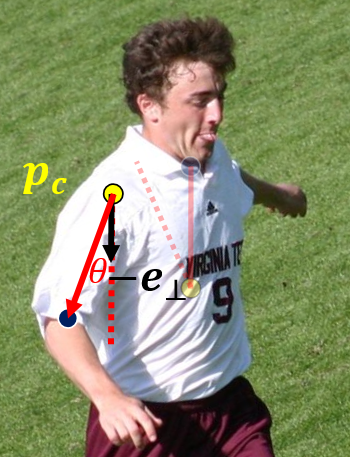}
		\vspace{-1mm}
		\caption{}
		\label{fig:s3figc}
	\end{subfigure}	
	\begin{subfigure}{0.11\textwidth} 
		\centering
		\includegraphics[width=0.945\linewidth]{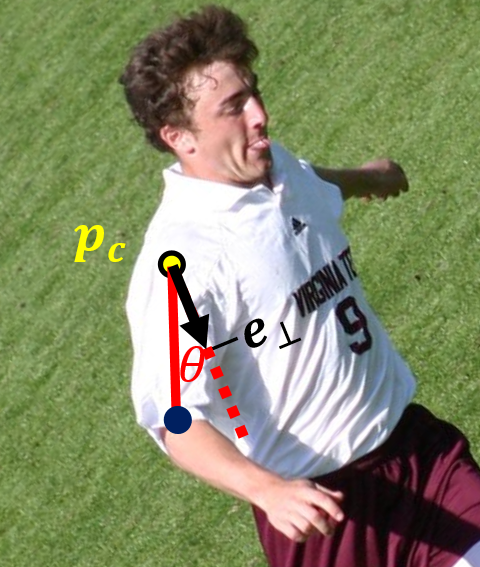}
		\vspace{-1mm}
		\caption{}
		\label{fig:s3figd}
	\end{subfigure}		
	\vspace{-1mm}
	\caption{Illustration of the body normalization and limb normalization. (a) and (b) show the rotation angle $\theta$ for body normalization. (c) is the image after body normalization and the rotation angle for a limb normalization. (d) shows the image after limb normalization. Note that our network actually performs the normalization on the score maps rather than on the image. From (b) to (d), we show the magnified view of the images for clarity.}
	\label{fig:angle}
\end{figure}

\vspace{0.1cm}
\noindent\textbf{Body normalization.} The purpose of body normalization is to make the orientation of the whole body the same, \eg, upright in our implementation\footnote[1]{Essentially, any orientation is fine in our approach.}. Specifically, we rotate the body as well as the $K$ score maps around the center of the four joints (\ie, left shoulder, right shoulder, left hip, right hip) so that the line from the center to the neck joint is upright, as shown in Figure~\ref{fig:angle} (b). The positions of the joints are estimated from the $K$ Gaussian-smoothed score maps by finding the maximum responses in each map and returning the corresponding position as the position of the joint.

We implement the normalization through spatial transform, which is written as follows,
\begin{align}
\bar{\mathbf{x}} = \mathbf{R} (\mathbf{x} - \mathbf{c}) + \mathbf{c},
\end{align}
where $\mathbf{c}$ is defined as the center of the four joints on the torso, $\mathbf{c} = \frac{1}{4}(\mathbf{p}_{l-shoulder} + \mathbf{p}_{r-shoulder} + \mathbf{p}_{l-hip} + \mathbf{p}_{r-hip})$, $\mathbf{p}_{l-shoulder}$ denotes the estimated location of the left shoulder joint,
and $\mathbf{R}$ is a rotation matrix,
\begin{equation}
\mathbf{R} = \begin{bmatrix}
\cos \theta & -\sin \theta \\[0.3em]
\sin \theta & \cos \theta
\end{bmatrix}.
\end{equation}
Here $\theta = \arccos \frac{(\mathbf{p}_{neck} - \mathbf{c})\cdot \mathbf{e}_{\perp}}{\|\mathbf{p}_{neck} - \mathbf{c}\|_2}$,  $\mathbf{e}_{\perp}$ denotes the unit vector along the vertical upward direction, which is illustrated in Figure~\ref{fig:angle}.

\begin{figure}[t]
	\vspace{1mm}
	\begin{subfigure} {0.156\textwidth}
		\centering
		\includegraphics[width=0.99\linewidth]{Original_relative_left_wrist_relative_to_shoulder_2000-green.pdf}
		\vspace{-5mm}
		\caption{}
		\label{fig:s4figa}
	\end{subfigure}
	\begin{subfigure}{0.156\textwidth}
		\centering
		\includegraphics[width=0.99\linewidth]{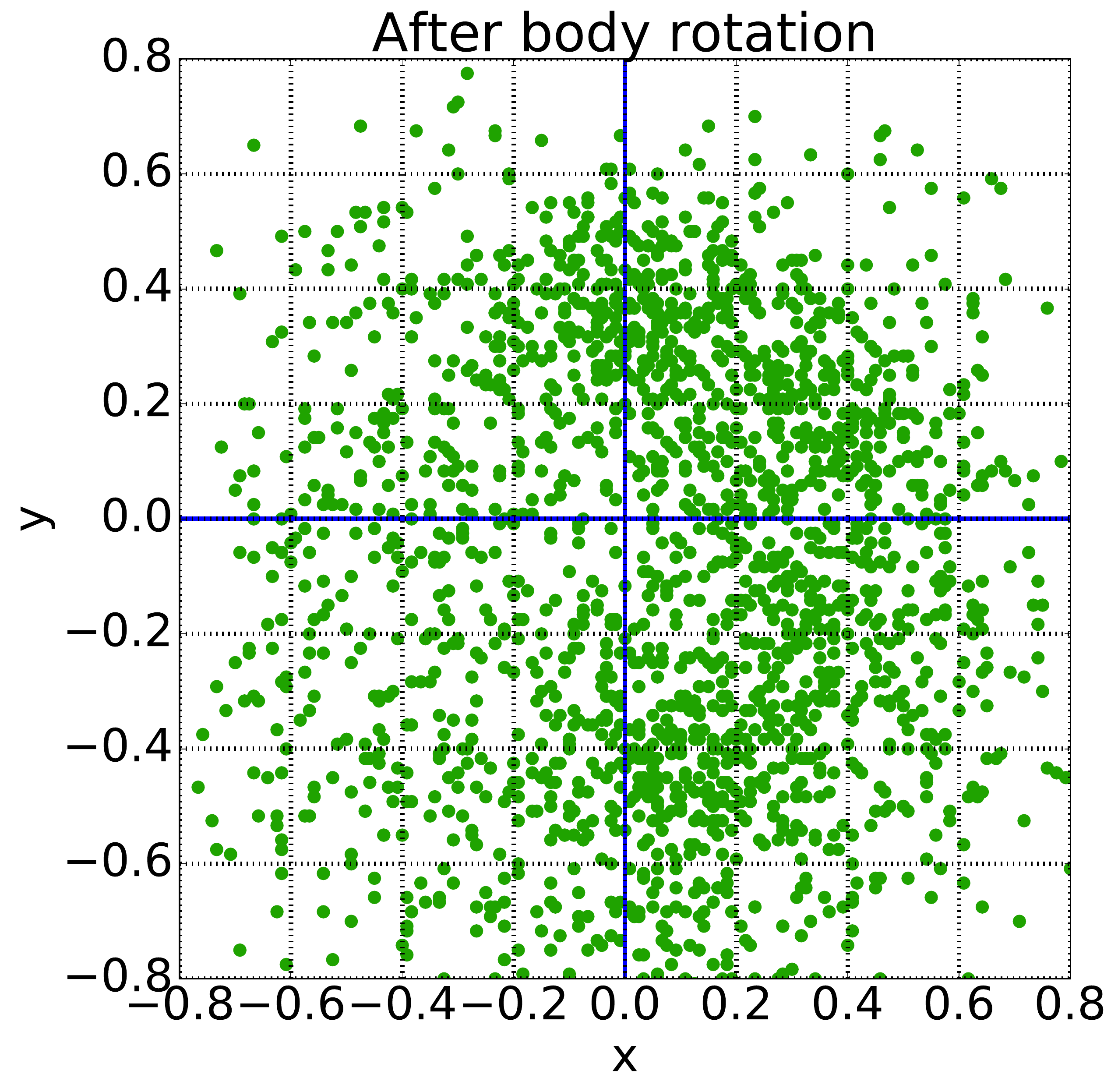}
		\vspace{-5mm}
		\caption{}
		\label{fig:s4figb}
	\end{subfigure}
	\begin{subfigure}{0.156\textwidth}
		\centering
		\includegraphics[width=0.99\linewidth]{Norm_Left_Wrist_2000-green.pdf}
		\vspace{-5mm}
		\caption{}
		\label{fig:s4figc}
	\end{subfigure}
	\vspace{-1mm}
	\caption{Limb normalization can compact the relative position distribution for some joints on limbs, which is hard to address via body normalization. (a), (b) and (c) are the relative positions of left wrist with respect to the position of left shoulder, and the relative positions after body normalization, and that after limb normalization. The distribution in (c) is much more compact.}
	\label{fig:distribution}
\end{figure}

\vspace{0.1cm}
\noindent\textbf{Local normalization.}
The end joints on the four limbs have higher variations. As illustrated in Figure~\ref{fig:distribution} (a) and (b), through body normalization, the distribution of the wrist with respect to the shoulder is still not compact. Limb normalization is then adopted where we rotate the arm to have upper arm vertical downwards, with the distribution, as shown in Figure \ref{fig:distribution} (c), becoming much more compact. There are four local normalization modules corresponding to the four limbs respectively. Each limb contains three joints: a root joint (shoulder, hip), a middle joint (elbow, knee), and an end joint (wrist, ankle). We perform the normalization by rotating the corresponding three score maps around the root joint such that the line connecting the root joint and the middle joint has a consistent orientation, \eg, vertical downwards in our implementation. The normalization process is illustrated in Figures~\ref{fig:angle} (c) and (d).

\vspace{0.1cm}
\noindent\textbf{Discussions.}
There are some alternative solutions for handling the diverse distribution problem of relative locations of the joints, \eg, type supervision~\cite{chu2016structure}, and mixture model~\cite{Yang2013Articu}. To check the effectiveness of our proposed normalization scheme, we provide two alternative solutions by considering the diversity of pose types. Here, we obtain pose type information by clustering human poses into three types from the LSP dataset. 

In our first alternative solution (\emph{type-supervision}), based on our global refinement framework, we remove the normalization model but add type supervision in the refinement network by learning three sets of score maps (\ie 3$\times$$K$+1) rather than one set. In the second alternative solution (\emph{multi-branch}), based on our global refinement framework, we remove the normalization model but extend the refinement network to multi-branches, with each branch handling one type of pose. Note that the number of parameters for the three-branch spatial configuration refinement is three times of ours. Specifically, for the alternative two solutions, we process the training data with extra data augmentation, to make the number of training data for each type similar to ours. The multi-branch approach is computationally more expensive and requires more training time than ours. 

We take the original FCN~\cite{long2015fully} as the joint detector, and make a comparison among the two alternative solutions and our global normalization refinement scheme. Table~\ref{tab:comparewithmixture} shows the results. The two alternative solutions improve performance over FCN, but under-performs our approach. It is possible to further improve the performance of the multi-branch approach with more extensive data augmentation and  more branches, but this will increase the computational complexity for both training and testing. In addition, our approach can also benefit from the multi-branch and type supervision solutions, where our normalization is applied to each branch and the type supervision further constrains the degrees of freedom of parts.

\begin{table}[t]
	\fontsize{7pt}{8pt}\selectfont\centering
	\tabcolsep=0.9pt
	\begin{center}
	\begin{tabular}{|c|ccccccc|c|c|}
		\hline
		& Head &	Shoulder &	Elbow	& Wrist	& Hip	& Knee	& Ankle &	Total & \#param.\\
		\hline\hline
		FCN & 93.3	& 86.7	& 74.4	&  68.0	&  85.7	&  82.0	&  78.5 & 	81.5   & 134M \\ %
		Type-supervision & 93.5	& 87.5	& 76.4	&  68.8	&  87.8	&  82.8	&  79.5 & 	82.3   & (134+11)M \\ %
		Multi-branch & \textbf{93.8}	& 87.7	& 76.3	&  69.4	&  87.7	&  82.6	&  79.8	&  82.5  & (134+33)M\\ %
		Global normalization & 93.7	& \textbf{88.8}	& \textbf{77.3}	& \textbf{69.6}	& \textbf{88.4}	& \textbf{84.0}	& \textbf{81.0}	& \textbf{83.3}  & (134+11)M\\ %
		\hline
	\end{tabular}
	\end{center}
	\vspace{-3mm}
	\caption{Comparing our global normalization-based solution with type supervision and the multi-branch solution on LSP dataset with the OC annotation (@PCK 0.2) trained on the LSP dataset.}
	\label{tab:comparewithmixture}
	\vspace{-2mm}
\end{table}

\begin{figure}[t!]
	\begin{subfigure} {0.11\textwidth}
		\centering
		\includegraphics[width=0.99\linewidth]{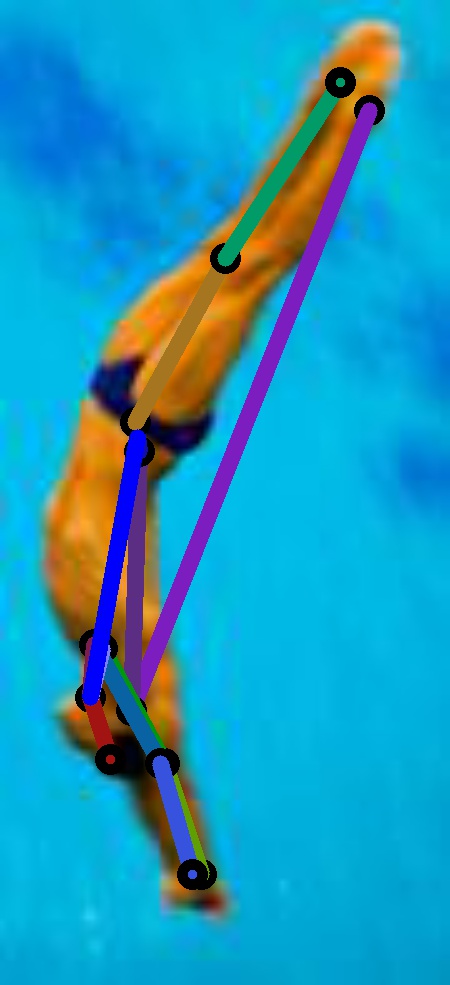}
		\vspace{-1mm}
	\end{subfigure}
	\begin{subfigure}{0.11\textwidth}
		\centering
		\includegraphics[width=0.99\linewidth]{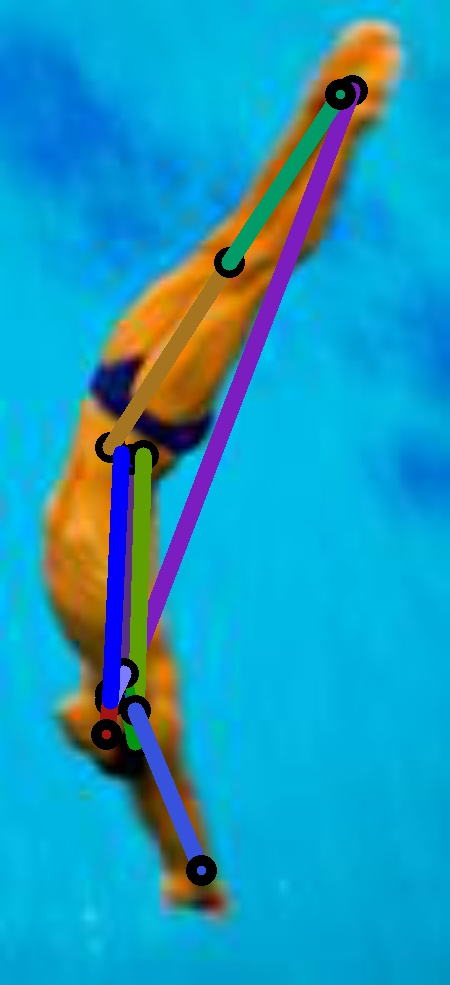} 
		\vspace{-1mm}
	\end{subfigure}
	\begin{subfigure}{0.11\textwidth}
		\centering
		\includegraphics[width=0.99\linewidth]{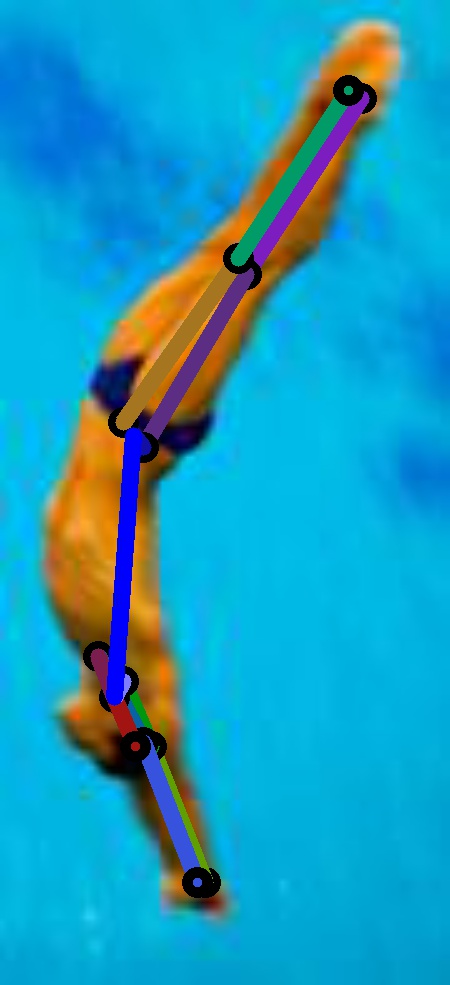}
		\vspace{-1mm}
	\end{subfigure}
	\begin{subfigure}{0.11\textwidth}
		\centering
		\includegraphics[width=0.99\linewidth]{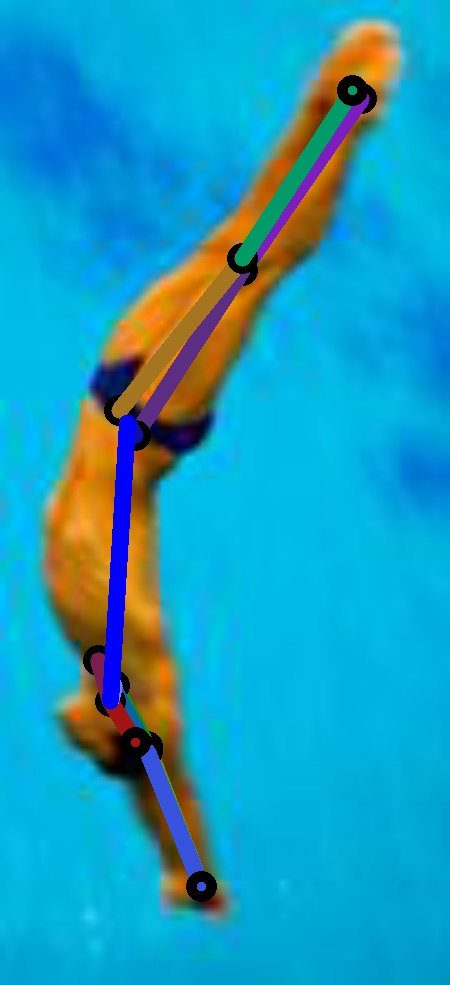}
		\vspace{-1mm}
	\end{subfigure}
	\begin{subfigure} {0.115\textwidth}
		\centering
		\includegraphics[width=0.99\linewidth]{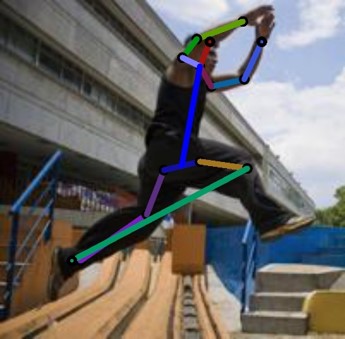} 
		\vspace{-4mm}
		\caption{}
		\label{fig:s5figa}
	\end{subfigure}
	\begin{subfigure}{0.115\textwidth}
		\centering
		\includegraphics[width=0.99\linewidth]{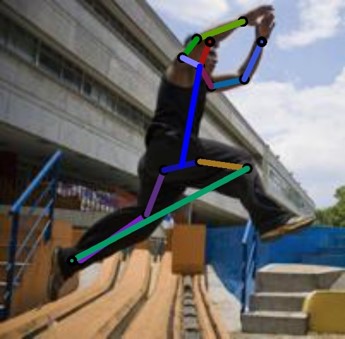} 
		\vspace{-4mm}
		\caption{}
		\label{fig:s5figb}
	\end{subfigure}	
	\begin{subfigure}{0.115\textwidth}
		\centering
		\includegraphics[width=0.99\linewidth]{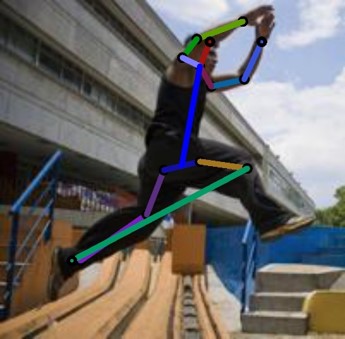} 
		\vspace{-4mm}
		\caption{}
		\label{fig:s5figc}
	\end{subfigure}
	\begin{subfigure}{0.115\textwidth}
		\centering
		\includegraphics[width=0.99\linewidth]{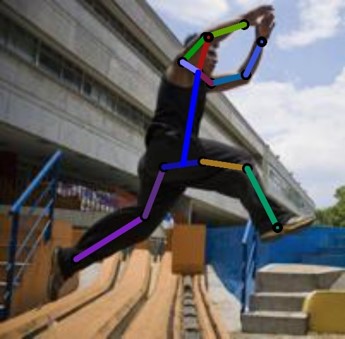} 
		\vspace{-4mm}
		\caption{}
		\label{fig:s5figd}
	\end{subfigure}
	\vspace{-2mm}	
	\caption{Estimated poses from (a) FCN, (b) the scheme with spatial refinement but without body normalization and limb normalization, (c) our scheme with body normalization, (d) our scheme with both body normalization and limb normalization.}
	\label{fig:stage-results}
\end{figure}

Figure \ref{fig:stage-results} shows examples of estimated poses from different stages of our network (Figure~\ref{fig:stage-results} (a), (c), (d)), and that from the similar network but without normalization modules (\ie, Figure~\ref{fig:stage-results} (b)). With global and local normalization, joint estimation accuracy is much improved (\eg, knee on the top images, ankle on the bottom images). Our normalization scheme can reduce the diversity of human poses and facilitate the inference process. For example, the consistent orientation of the left hip and the left knee makes it easier to infer the location of the left ankle on the bottom example in Figure \ref{fig:stage-results}. 

\subsection{Multi-Scale Supervision and Fusion for Joint Detection}
To efficiently train the FCN and exploit intermediate-level representations, we introduce multi-scale supervision and multi-scale fusion, which show performance gain in many works~\cite{Lee2015, long2015fully, szegedy2015googlenet,  xie2015holistically}. The network structure is provided in Figure~\ref{fig:FCNN}. Multi-scale supervision makes the network concentrate on accurate localization on different resolutions, avoiding loss in accuracy due to down-sampling. This is different from \cite{hourglass16, wei2016cpm}, adding multi-supervision to each stage with the same resolution. Multi-scale fusion exploits the information at different scales. More details are introduced in the Section \ref{implementation}.

\subsection{Implementation Details}

\noindent\textbf{Network architectures.} The proposed network architecture contains three main parts: the base network for joint detection, the normalization network, and the refinement network.

\begin{figure}[t!]
	\centering
	\includegraphics[width=0.96\linewidth]{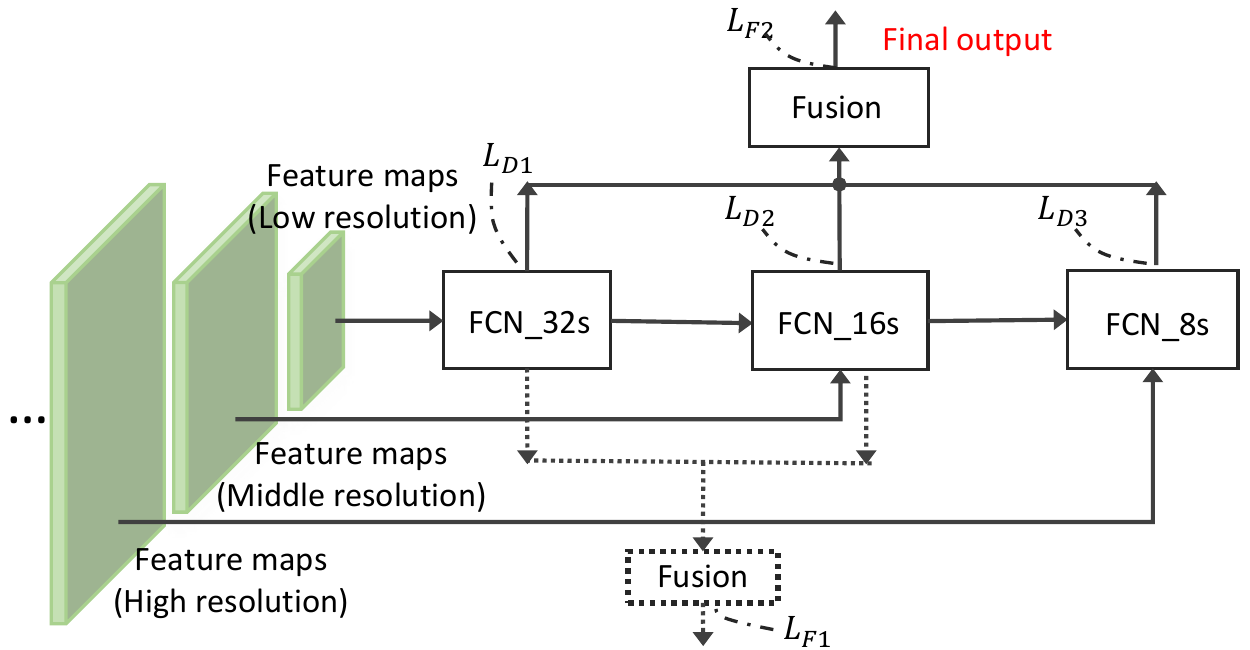}
	\caption{Architecture of the improved FCN. We utilize multi-scale supervision and fusion.}
	\label{fig:FCNN}
\end{figure}

\emph{Joint detection network}: We use fully convolutional network as our joint detector. For fairness of comparison, we use architectures similar to the compared methods as the joint detectors. We demonstrate the effectiveness of our normalization scheme on top of different joint detectors: the improved FCN as showed in Figure~\ref{fig:FCNN}, ResNet-152 similar to ~\cite{insafutdinov16eccv,regression2016pose} and Hourglass~\cite{hourglass16}. The FCN generates three sets of score maps (FCN$\_$32s, FCN$\_$16s, and FCN$\_$8s), at different resolutions, corresponding to the last three deconvolution layers with strides 32, 16, and 8 respectively. The fusion is an ensemble of different scales with a 1$\times$1 convolutional layer. We introduce multi-scale supervision (with losses $L_{D1}$, $L_{D2}$, and $L_{D3}$) and multi-scale fusion (with losses $L_{F1}$, $L_{F2}$) (see Figure~\ref{fig:FCNN}). The architecture of Hourglass is the same as ~\cite{hourglass16} and ResNet is similar to ~\cite{insafutdinov16eccv}. 

\begin{figure}[t!]
	\centering
	\includegraphics[width=0.95\linewidth]{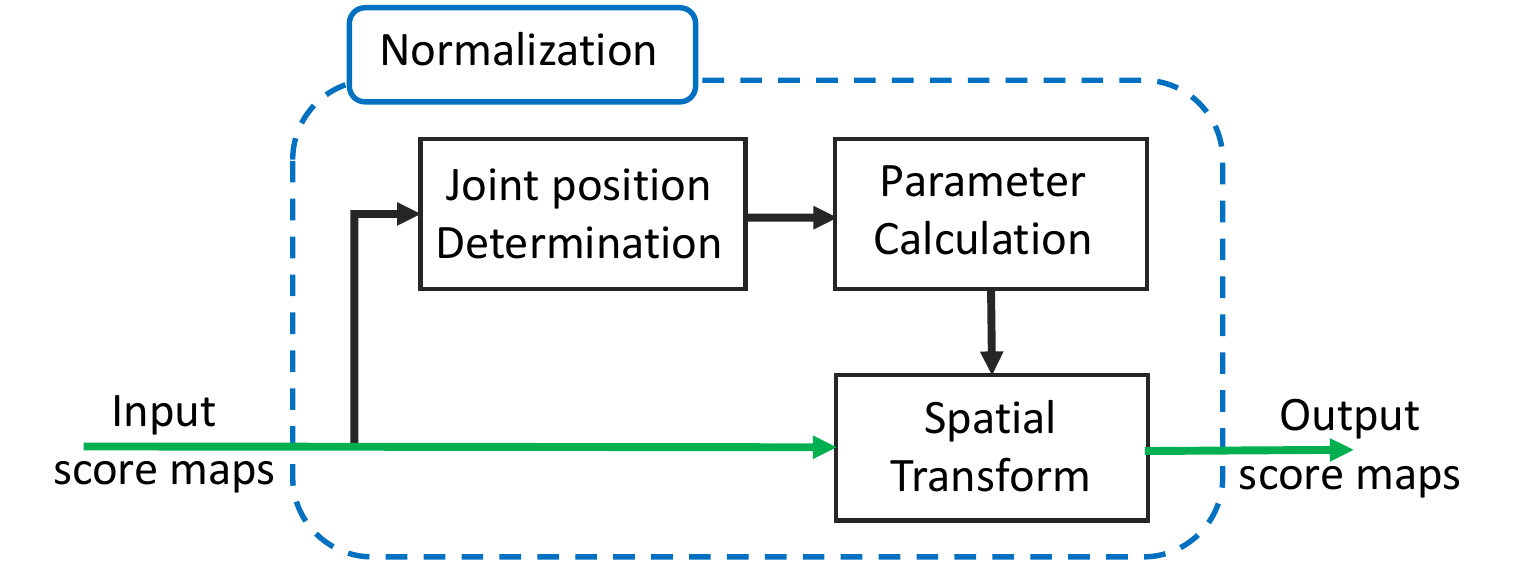}
	\vspace{-1mm}
	\caption{Normalization module. Based on the score maps, a module derives the joint positions. Then the spatial transform parameters can be calculated directly.}
	\label{fig:normalization}
	\vspace{-3mm}
\end{figure}

\emph{Normalization network}: In Figure~\ref{fig:normalization}, we show the flowchart of the normalization module. The spatial transform is performed on the score maps with the calculated transform parameters. End-to-end training is supported with the  error back propagation along the transform path (as denoted by the green line). For the joint position determination module, a Gaussian blur is performed on the mapped score maps, with the mapping corresponding to the Sigmoid like operation or no operation, depending the loss design of the joint detection network. Then, the position corresponding to the maximal value in each processed score map is estimated as the position of that joint. The network calculates the rotation center $\mathbf{c}$ and the rotation angle $\theta$ based on the estimated positions of joints. All the operations are incorporated into the network as layers.

\emph{Refinement network}: The refinement network consists of four convolutional layers. The convolutional kernel sizes and channel numbers for the four layers are $9\times9\times (K+1)$ with 128 output channels, $15\times15\times 128$ with 128 output channels, $15\times15\times 128$ with 128 output channels, and  $1\times1\times 128$ with $J$ output channels, where $J$ denotes the number of output joints. Large kernel sizes of 9$\times$9 and 15$\times$15 are beneficial for capturing spatial information of each joint. 


\vspace{0.1cm}
\noindent\textbf{Loss functions.} The groundtruth is generated to be $K$ score maps. When FCN or ResNet detector is utilized, the pixels in a circled region with radius of $r$ centered at a joint are labeled by 1 while other pixels are set by 0. We define the radius $r$ as 0.15 times of the distance between left shoulder and right hip. When Hourglass detector is utilized, 2D Gaussian centered on the joint location is used for groundtruth labeling \cite{hourglass16}. For the spatial refinement stages, both visible and occluded joints are labeled.   

For the FCN joint detector, softmax function is utilized to estimate the probability of being some joint for the visible joints. For spatial refinement, after the several convolution layers, sigmoid-like operation $1/(1+e ^{-(wx+b)})$ is used to map the score $x$ to the estimation of the probability of being some joint (both visible and occluded joints). Here, $w$ and $b$ are two learnable parameters which transform the scores to be in a suitable range of the domain of sigmoid function. 
For other joint detectors, \ie ResNet and Hourglass, we use their designed loss.

\vspace{.1cm}
\noindent\textbf{Optimization.}
We pretrain the network by optimizing joint detector, global and local refinement model, and then fine-tune the whole framework. 

We initialize the parameters of the refinement model randomly with a Gaussian distributed variable of variance 0.001. When the network of joint detector converges, we fix it and train the body refinement network with a base learning rate of 0.001. Afterwards, we fix the former networks and train the limb refinement network. Finally, we fine-tune the entire network with learning rate 0.0002.

For FCN, we initialize it with the model weights from PASCAL VOC \cite{long2015fully}. During training, we progressively minimize the loss function:
first minimize $L_{D1}$
then $L_{D1} + L_{D2} + L_{F1}$,
then  $L_{D1} + L_{D2} + L_{D3} + L_{F1} + L_{F2}$ (see Figure \ref{fig:FCNN}). FCN detector is implemented based on Caffe and SGD is taken as the optimization algorithm. The initial learning rate is set to 0.001. For other joint detectors, such as ResNet-152 and Hourglass, we adopt the same settings as proposed by the authors in their papers.

\label{implementation}

\section{Experiments}
\noindent\textbf{Datasets.}
We evaluate the proposed method on four datasets: Leeds Sports Pose (LSP) \cite{Johnson2010LSP}, extended LSP (LSPET) \cite{Johnson2011LSP}, Frames Labeled in Cinema (FLIC) \cite{Sapp13FLIC} and MPII Human Pose \cite{mpiidataset}. The LSP dataset contains 1000 training and 1000 testing images from sports activities, with 14 full body joints annotated. The LSPET dataset adds 10,000 more training samples to the LSP dataset. The FLIC dataset contains 3987 training and 1016 testing images with 10 upper body joints annotated. The MPII Human Pose dataset includes about 25k images with 40k annotated poses. Existing works evaluate the performance on the LSP dataset with different training data, which we follow for performance comparisons respectively.


\noindent\textbf{Evaluation criteria.}
The metrics ``Percentage of Correct Keypoints (PCK)" and the ``Area Under Curve (AUC)" are utilized for evaluation \cite{Yang2013Articu, pishchulin16cvpr}. 
A joint is correct if it falls within $\alpha\cdot l_r$ pixels of the groundtruth position, with $\alpha$ denoting a threshold and $l_r$ a reference length. $l_r$ is the torso length for the LSP, FLIC, and the head size for the MPII. 




\noindent\textbf{Data Augmentation.}
For the LSP dataset, we augment the training data by performing random scaling with a scaling factor between 0.80 and 1.25, horizontal flipping, and  rotating the data across 360 degrees, in consideration of its unbalanced distribution of pose orientations. All input images are resized to $340\times340$ pixels. For the FLIC and the MPII dataset, we randomly rotate the data across +/- 30 degrees and resize images into $256\times256$ pixels.

\begin{table}[b] 
	\vspace{-2mm}
	\fontsize{7pt}{8pt}\selectfont\centering 
	\tabcolsep=1pt
	\begin{tabular}{|c|ccccccc|c|c|}
		\hline
		& Head  & Shoulder & Elbow & Wrist & Hip   & Knee  & Ankle & Total & AUC\\
		\hline\hline
		Kiefel et al. \cite{kiefel2014human} & 83.5  & 73.7  & 55.9  & 36.2  & 73.7  & 70.5  & 66.9  & 65.8  & 38.6\bigstrut\\
		Ramakrishna et al. \cite{Ramakrishna2014posemachines} & 84.9  & 77.8  & 61.4  & 47.2  & 73.6  & 69.1  & 68.8  & 69.0    & 35.2  \bigstrut\\
		Pishchulin et al. \cite{pishchulin2013poselet}& 87.5  & 77.6  & 61.4  & 47.6  & 79.0    & 75.2  & 68.4  & 71.0    & 45.0\bigstrut\\
		Ouyang et al. \cite{ouyang2014multi} & 86.5  & 78.2  & 61.7  & 49.3  & 76.9  & 70.0    & 67.6  & 70.0    & 43.1 \bigstrut\\
		Chen\&Yuille \cite{Chen_NIPS14} & 91.5  & 84.7  & 70.3  & 63.2  & 82.7  & 78.1  & 72.0    & 77.5  & 44.8 \bigstrut\\
		Yang et al. \cite{yang2016end} & 90.6  & \textbf{89.1} & \textbf{80.3} & 73.5  & 85.5  & 82.8  & 68.8  & 81.5  & 43.4 \bigstrut\\
		Chu et al. \cite{chu2016structure} & 93.7  & 87.2  & 78.2  & \textbf{73.8} & 88.2  & 83.0    & \textbf{80.9} & 83.6  & 50.3  \bigstrut\\
		\hline
		Ours(FCN) & 94.3  & 87.8  & 77.1  & 69.8  & 87.1  & 83.7  & 79.7  & 82.8  & 54.2 \bigstrut\\
		Ours(FCN+Refine) & \textbf{94.9} & 88.8  & 77.6  & 70.7  & \textbf{88.9} & \textbf{84.8} & 80.5  & \textbf{83.7} & \textbf{54.6} \bigstrut\\
		\hline
	\end{tabular}%
	\vspace{-1mm}
	\caption{Performance comparison on the LSP testing set with the OC annotation (@PCK0.2) trained on the LSP training set.}
	\label{tab:LSP-OC}%
\end{table}%

\begin{table}[b] 
	\fontsize{7pt}{8pt}\selectfont\centering
	\tabcolsep=1.5pt
	\begin{tabular}{|c|ccccccc|c|c|}
		\hline
		& Head  & Shoulder & Elbow & Wrist & Hip   & Knee  & Ankle & Total & AUC\\
		\hline\hline
		Pishchulin et al. \cite{pishchulin16cvpr}& \textbf{97.4}  & \textbf{92.0}  & 83.8  & \textbf{79.0}  & 93.1  & 88.3  & 83.7  & 88.2    & \textbf{65.0}\bigstrut\\
		\hline
		Ours(FCN) & 96.2  & 90.7  & 83.3  & 77.5  & 91.2  & 89.3  & 85.0  & 87.6  & 61.8 \bigstrut\\
		Ours(FCN+Refine) & 96.7 & 91.8  & \textbf{84.4}  & 78.3  & \textbf{93.3} & \textbf{90.7} & \textbf{85.8}  & \textbf{88.7} & 63.0 \bigstrut\\
		\hline
	\end{tabular}%
	\vspace{-1mm}
	\caption{Performance comparison on the LSP testing set with the OC annotation (@PCK0.2) trained on the MPII+LSPET+LSP training set.}
	\label{tab:LSP-MPII-OC}%
\end{table}%

\subsection{Results} 

We denote our final model as \emph{Ours}\emph{(detector+Refine)}, and the scheme with only detector as \emph{Ours}\emph{(detector)}. All the experiments are conducted without any post-processing.

\vspace{0.1cm}
\noindent\textbf{LSP OC.}
With the LSP dataset as training data, Table \ref{tab:LSP-OC} shows the comparisons with the OC annotation for per-joint PCK results, the overall results at threshould $\alpha=0.2$ (@PCK0.2), and the AUC. Our method achieves the best performance, where the AUC is 4.3\% higher than Chu \etal \cite{chu2016structure}, even though the layer number of their additional network (180 conv layers) is much larger than our refinement network (24 conv layers).  Our refinement provides 1\% improvement in the overall accuracy. 

Another work \cite{pishchulin16cvpr} incorporates the MPII and LSPET dataset for training. The results with the same training set are shown in Table \ref{tab:LSP-MPII-OC}. Our refinement achieves 1.1\% improvement in overall accuracy and outperforms the start-of-the-art even though our detector does not use location refinement and an auxiliary task as used by~\cite{pishchulin16cvpr}.

\noindent\textbf{LSP PC.} Table \ref{tab:LSP-PC-all} shows the comparisons with PC annotation. Compared with the result of Yang \etal \cite{yang2016end} on the LSP dataset, our method significantly improves the performance by 4.3\% in overall accuracy and 11.4\% in AUC.



\begin{table}[t] 
	\fontsize{7pt}{8pt}\selectfont\centering 
	\tabcolsep=1.5pt
	\begin{tabular}{|c|ccccccc|c|c|}
		\hline
		& Head  & Shoulder & Elbow & Wrist & Hip   & Knee  & Ankle & Total   & AUC  \bigstrut\\
		\hline\hline
		Tompson et al. \cite{tompson2014joint} & 90.6  & 79.2  & 67.9  & 63.4  & 69.5  & 71.0    & 64.2  & 72.3  & 47.3  \bigstrut\\
		Fan et al. \cite{fan2015combing} & 92.4  & 75.2  & 65.3  & 64.0    & 75.7  & 68.3  & 70.4  & 73.0    & 43.2   \bigstrut\\
		Carreira et al. \cite{IEF16} & 90.5  & 81.8  & 65.8  & 59.8  & 81.6  & 70.6  & 62.0    & 73.1  & 41.5   \bigstrut\\
		Chen\&Yuille \cite{Chen_NIPS14} & 91.8  & 78.2 & \textbf{71.8} & 65.5  & 73.3  & 70.2  & 63.4  & 73.4  & 40.1  \bigstrut\\
		Yang et al. \cite{yang2016end} & 90.6  & 78.1  & 73.8  & \textbf{68.8} & 74.8  & 69.9  & 58.9 & 73.6  & 39.3  \bigstrut\\
		\hline
		Ours(FCN) & 93.8  & 80.3  & 69.7  & 64.7  & 81.0    & 78.1  & 73.1  & 77.2  & 50.5 \bigstrut\\
		Ours(FCN+Refine) & \textbf{94.0} & \textbf{80.9}  & 70.6  & 65.3  & \textbf{82.3} & \textbf{78.5} & \textbf{73.7}  & \textbf{77.9} & \textbf{50.7} \bigstrut\\
		\hline
	\end{tabular}%
	\vspace{-1mm}
	\caption{Performance comparison on the LSP testing set with PC (@PCK0.2) trained on the LSP training set.}
	\label{tab:LSP-PC-all}%
\end{table}%

\begin{table}[t] 
	\fontsize{7pt}{8pt}\selectfont\centering 
	\tabcolsep=1.5pt
	\begin{tabular}{|c|ccccccc|c|c|}
		\hline
		& Head  & Shoulder & Elbow & Wrist & Hip   & Knee  & Ankle & Total   & AUC  \bigstrut\\
		\hline\hline
		Bulat et al. \cite{regression2016pose} & \textbf{98.4} & 86.6  & 79.5  & 73.5    & 88.1  & 83.2  & 78.5  & 83.5  & -- \bigstrut\\
		Wei et al. \cite{wei2016cpm} & -- & -- & -- & --    & --  & --  & --  & 84.32  & -- \bigstrut\\
		Rafi et al. \cite{BMVC16pose} & 95.8 & 86.2  & 79.3  & 75.0    & 86.6  & 83.8  & 79.8  & 83.8  & 56.9  \bigstrut\\
		Yu et al. \cite{yu2016ECCV16Basis}& 87.2  & 88.2  & \textbf{82.4} & \textbf{76.3} & \textbf{91.4} & \textbf{85.8} & 78.7  & 84.3  & 55.2  \bigstrut\\
		\hline
		Ours(FCN) & 95.2  & 86.2  & 78.1  & 72.8  & 87.0    & 85.7  & 81.3  & 83.7  & 56.1  \bigstrut\\
		Ours(FCN+Refine) & 95.5  & \textbf{88.5} & 80.0    & 73.9  & 89.8  & \textbf{85.8} & \textbf{81.5} & \textbf{85.0} & \textbf{58.5} \bigstrut\\
		\hline
	\end{tabular}%
	\vspace{-1mm}
	\caption{Performance comparison on the LSP testing set with PC (@PCK0.2) trained on the LSP+LSPET training set.}
	\label{tab:LSPET-PC}%
	\vspace{-3mm}
\end{table}%

\vspace{0.1cm}
We incorporate the LSPET dataset into the training data and evaluate the performance with PC annotation. From Table \ref{tab:LSPET-PC}, we can see that our scheme achieves the best performance.  
Yu \etal \cite{yu2016ECCV16Basis} extracts many pose bases to represent various human poses. In contrast, our method normalizes various poses. Our method outperforms theirs by 3.3\% in AUC and 0.7\% in the overall accuracy. 

To verify the effectiveness of our normalization scheme, we connect our refinement model at the end of those deeper joint detectors, \ie, ResNet-152 (152 layers)~\cite{insafutdinov16eccv,regression2016pose} and Hourglass (about 300 layers)~\cite{hourglass16}. The results are shown in Table \ref{tab:LSP-MPII-PC}. Without using the location refinement and auxiliary task ~\cite{insafutdinov16eccv}, our baseline scheme \emph{Ours(ResNet-152)} drops about 1\% than~\cite{insafutdinov16eccv}. With the proposed refinement added, our scheme \emph{Ours(ResNet+Refine)} improves over the baseline by 1\% in the overall accuracy and is comparable to \cite{insafutdinov16eccv}. Bulat \etal ~\cite{regression2016pose} added a modified hourglass network ~\cite{hourglass16} (90 layers, with parameters being three times larger than our refinement model) after ResNet-152. \emph{Ours(ResNet+Refine)} is 1.4\% better than that of Bulat \etal ~\cite{regression2016pose} in AUC. When we take Hourglass as our detector, the proposed refinement brings 0.5\% improvement in the overall accuracy.

\begin{table}[t] 
	\fontsize{7pt}{8pt}\selectfont\centering 
	\tabcolsep=0.8pt
	\begin{tabular}{|c|ccccccc|c|c|}
		\hline
		& Head  & Shoulder & Elbow & Wrist & Hip   & Knee  & Ankle & Total   & AUC \bigstrut\\
		\hline\hline
		Insafutdinov et al. \cite{insafutdinov16eccv} & 97.4  & 92.7  & 87.5  & 84.4  & 91.5  & 89.9  & 87.2  & 90.1  & \textbf{66.1} \bigstrut\\
		Wei et al. \cite{wei2016cpm} & 97.8  & 92.5  & 87.0    & 83.9  & 91.5  & 90.8  & 89.9  & 90.5  & 65.4 \bigstrut\\
		Bulat et al. \cite{regression2016pose} & 97.2  & 92.1  & 88.1  & 85.2  & 92.2  & \textbf{91.4}  & 88.7  & 90.7  & 63.4  \bigstrut\\
		\hline
		Ours(ResNet-152) & 97.0    & 91.5  & 86.2  & 82.8  & 89.4  & 89.9  & 87.5  & 89.2  & 63.5 \bigstrut\\
		Ours(ResNet+Refine) & 97.3	& 92.2 &	87.1&	83.5&	92.1&	90.6&	87.8&	90.1&	64.8 \bigstrut\\
		Ours(Hourglass) & 97.7	& 93.0 &	88.3&	84.8&	92.3&	90.2&	90.0&	90.9&	65 \bigstrut\\
		Ours(Hg+Refine) & \textbf{97.9} & \textbf{93.6} & \textbf{89.0} & \textbf{85.8} & \textbf{92.9} & 91.2 & \textbf{90.5} & \textbf{91.6}	& 65.9 \bigstrut\\
		\hline
	\end{tabular}%
	\vspace{-1mm}
	\caption{Performance comparison on the LSP testing set with the PC annotation (@PCK0.2) trained on the MPII+LSPET+LSP training set.}
	\label{tab:LSP-MPII-PC}%
\end{table}%

\vspace{0.1cm}
\noindent\textbf{FLIC dataset.} We evaluate our method on the FLIC dataset with the OC annotation. We take ResNet-152 \cite{he2016Resi} as our joint detection network. Table \ref{tab:FLIC} shows that our refinement improves over the baseline model by 0.4\% for elbow, 0.7\% for wrist, and 1.6\% in AUC.

\begin{table}[t!]
\fontsize{7pt}{8pt}\selectfont\centering 
\tabcolsep=7pt
\begin{tabular}{|c|cccc|c|}
	\hline
	& Head & Shoulder & Elbow & Wrist & AUC \bigstrut\\
	\hline\hline
	Toshev et al. \cite{toshev2014deeppose} & --  & --  & 92.3 & 82 & -- \bigstrut\\
	Tompson et al. \cite{tompson2014joint} & --  & --  & 93.1 & 89 & -- \bigstrut\\
	Chen\&Yuille.\cite{Chen_NIPS14}& --  & --  & 95.3 & 92.4 & -- \bigstrut\\
	Wei et al. \cite{wei2016cpm} & --  & --  & 97.6 & 95 & -- \bigstrut\\
	Newell et al. \cite{hourglass16}& --  & --  & 99.0 & 97.0 & -- \bigstrut\\
	\hline
	ResNet-152 & 99.7 & 99.7 & 99.1 & 97 & 75.3 \bigstrut\\
	Ours(Refine) & \textbf{99.9} & \textbf{99.8} & \textbf{99.5} & \textbf{97.7} & \textbf{76.9} \bigstrut\\
	\hline
\end{tabular}%
\vspace{-1mm}
\caption{Performance comparison on the FLIC dataset with OC annotation (@PCK0.2).}
\label{tab:FLIC}%
\end{table}%


\begin{table}[t]
	\fontsize{7pt}{8pt}\selectfont\centering
	\tabcolsep=1.6pt
	\begin{tabular}{|c|ccccccc|c|}
		\hline
		& Head &	Shoulder &	Elbow	& Wrist	& Hip	& Knee	& Ankle &	Total \\
		\hline\hline
		Hourglass [22] & \textbf{98.2} & \textbf{96.3} & \textbf{91.2} & 87.1 & \textbf{90.1} & \textbf{87.4} & 83.6 & 90.9 \\
		Ours(Hourglass+Refine) & 98.1 & 96.2 & \textbf{91.2} & \textbf{87.2} & 89.8 & \textbf{87.4} & \textbf{84.1} & \textbf{91.0} \\
		\hline
	\end{tabular}
	\vspace{-1mm}
	\caption{Performance comparison on the MPII test set (@PCKh0.5) trained on the MPII training set.}
	\label{tab:hg-MPII-test}
	\vspace{-3mm}
\end{table}

\begin{figure}[t] 
\vspace{-4mm}
	\begin{subfigure}[t]{0.96\linewidth}
	\hspace{-4mm}
	\includegraphics[width=1.1\linewidth]{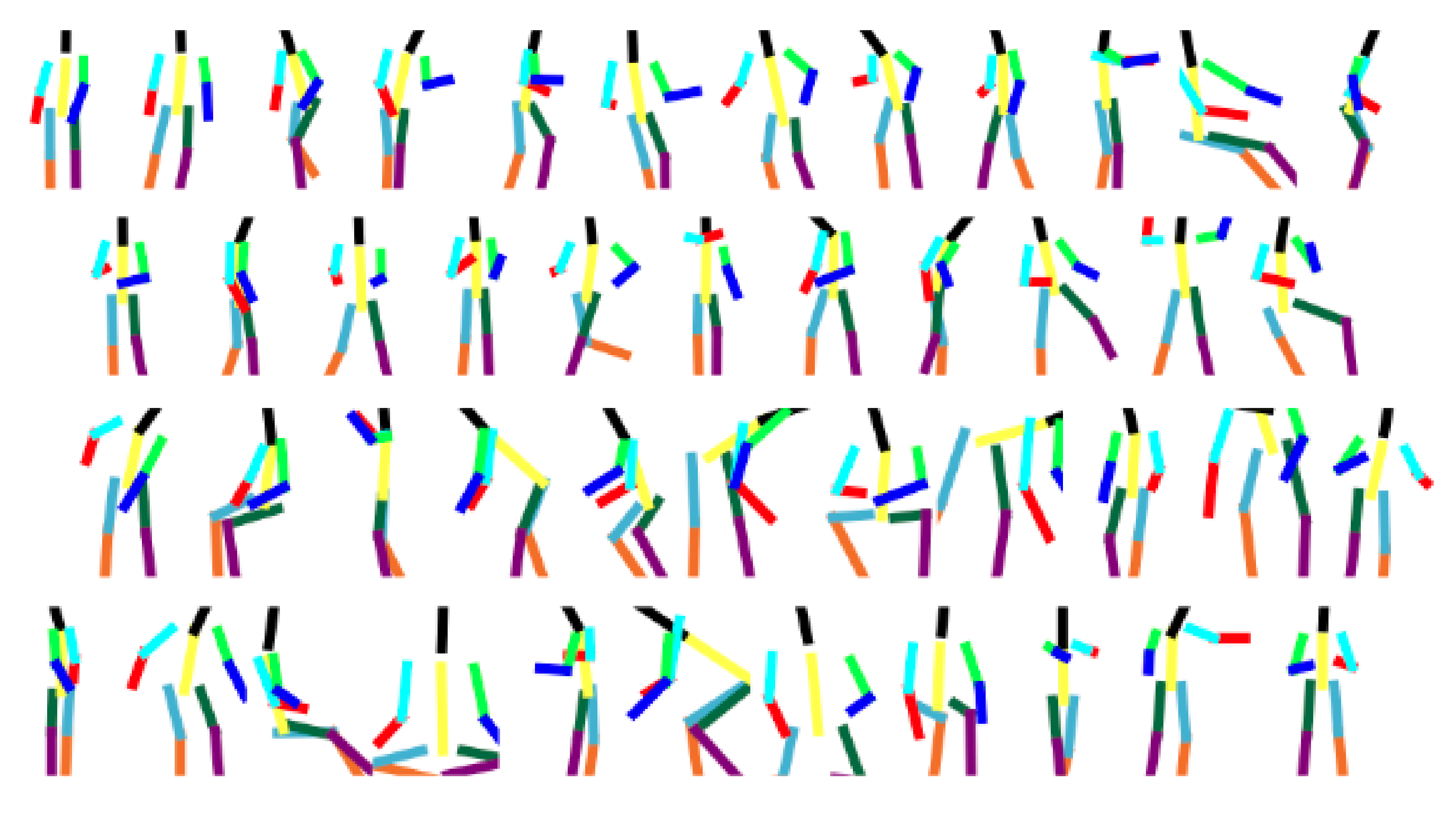}
	\end{subfigure}	
	\vspace{-3mm}
\caption[]{Body pose clusters on the MPII test set. The maker of MPII dataset clusters body poses into 45 types on the test set. Note the figure is from  \url{http://human-pose.mpi-inf.mpg.de./\#results}.} \label{fig:mpii-test-distribution}
\end{figure}

\vspace{0.1cm}
\noindent\textbf{MPII dataset.} We take Hourglass~\cite{hourglass16} as our joint detector and  evaluate our method on the MPII dataset. Table \ref{tab:hg-MPII-test} shows that our refinement performs similarly on the test set in overall accuracy. On the validation set, we obtains 0.4\% improvement. To check the reason for small gains, we analyze the relative position distribution on the MPII validation dataset. We found that the original distribution without normalization is already compact, being similar to the distribution after the normalization on the LSP dataset. Unlike the poses in the LSP dataset (sport poses), the majority of poses are upright and normal, as shown in Figure \ref{fig:mpii-test-distribution}. Our normalization scheme presents its advantages on the datasets including high diverse poses. In reality, these complicated postures are inevitable.

\begin{table}[t]
	\fontsize{7pt}{8pt}\selectfont\centering 
	\tabcolsep=2pt
	\begin{tabular}{|c|ccccccc|c|}
		\hline
		& Head & {Shoulder} & Elbow & Wrist & {Hip} & Knee & Ankle & {Total} \bigstrut\\
		\hline\hline
		FCN & \textbf{94.3} & {87.8} & 77.1 & 69.8 & {87.1} & 83.7 & 79.7 & {82.8} \bigstrut\\
		Stage-1 w/o body norm. & 93.5 & {88.2} & 77.2 & 69.8 & {87.5} & 83.8 & 80.2 & {82.9} \bigstrut\\
		Stage-1 w body norm. & 94.2 & {\textbf{88.5}} & \textbf{77.8} & 69.8 & 88.2 & 83.8 & 80.2 & 83.2 \bigstrut\\
		Stage-2 w/o limb norm. & 94 & 88.3 & 77.7 & {69.4} & 88 & 83.8 & {80.3} & 83 \bigstrut\\
		Stage-2 w limb norm. & 94.2 & 88.4 & 77.7 & \textbf{70.4} & \textbf{88.8} & \textbf{84.7} & \textbf{80.5} & \textbf{83.5} \bigstrut\\
		\hline
	\end{tabular}%
	\vspace{-1mm}
	\caption{Evaluation of body normalization and limb normalization on the LSP test dataset with the OC annotation (@PCK0.2) trained on the LSP training dataset.}
	\label{tab:poseNormalization}%
\end{table}%

\begin{table}[t]
	\fontsize{7pt}{8pt}\selectfont\centering 
	\tabcolsep=2pt
	\begin{tabular}{|c|ccccccc|c|}
		\hline
		& Head & {Shoulder} & Elbow & Wrist & {Hip} & Knee & Ankle & {Total} \bigstrut\\
		\hline\hline
		FCN & 94.9 & 90.5 & 82.2 & 74.8 & 89 & 88.2 & 83.5 & 86.2  \bigstrut\\
		Stage-1 w/o body norm. & 94.8 & 90.1 & 81.6 & 75.2 & 90 & 87.7 & 83 & 86  \bigstrut\\
		Stage-1 w body norm. & 94.9 & 90.8 & 83.8 & 76.3 & 89.7 & 88.3 & 84 & 86.8  \bigstrut\\
		Stage-2 w/o limb norm. & 95 & 88.3 & 80.6 & 75.7 & 88.5 & 86.2 & 82.8 & 85.3  \bigstrut\\
		Stage-2 w limb norm. & \textbf{95.4} & \textbf{91.1} & \textbf{84} & \textbf{76.8} & \textbf{90.9} & \textbf{89} & \textbf{84.4} & \textbf{87.4}  \bigstrut\\
		\hline
	\end{tabular}%
	\vspace{-1mm}
	\caption{Evaluation of body normalization and limb normalization on the LSP test dataset with the OC annotation (@PCK0.2) trained on the LSP+LSPET training dataset.}
	\label{tab:poseNormalizationLSPET}%
\end{table}%

\subsection{Ablation Study} 
We analyze the effectiveness of the proposed components, including the two pose normalization and refinement stages, and the multi-scale supervision and fusion.

\noindent\textbf{Global and local normalization.} 
To verify the effectiveness of body and limb normalization, we compare the results of the network with normalization versus that without normalization on the two stages separately.

Table \ref{tab:poseNormalization} shows the comparisons on the LSP dataset. With body normalization, the shoulder is 0.7\% higher than that of FCN and the hip estimation is improved by 1.1\%. In contrast, the model without the body normalization introduces much smaller improvement. With limb normalization, the accuracy of wrist, knee, and ankle is improved by 0.6\%, 0.9\%, and 0.3\% respectively. Without pose normalization, the subnetwork tends to preserve the results of the former stage. Similar phenomena are observed when we use the LSP+LSPET dataset for training as shown in Table \ref{tab:poseNormalizationLSPET}. We notice that the performance of Stage-1 without body normalization even provides interior performance than FCN. In contrast, when body normalization is utilized, consistent performance improvement can be achieved.

\noindent\textbf{Multi-scale supervision and fusion.} 
For FCN, we add multi-scale supervision and  multi-scale score map fusion to improve accuracy. Here, we evaluate the efficiency of the extra supervision and fusion respectively. Table \ref{tab:supervision} shows the experiment results. FCN$\_$16s and FCN$\_$8s denote the results of the original FCN without extra loss and fusion at the middle and high resolution respectively. 
FCN$\_$16s~(Extra) and FNC$\_$8s~(Extra) denote the results after adding supervision.
FCN$\_$16s~(Fusion) and FCN$\_$8s~(Fusion) denote the results after adding both supervision and fusion. From Table \ref{tab:supervision}, we have the following two observations. First, with extra supervision, the accuracy of most joints improves by more than 1\% and the AUC increases noticeably at the same resolution level.  Note that FCN$\_$8s~(Extra) achieves similar accuracy as FCN$\_$16s~(Extra) but its AUC is much higher. Second, we fuse the score maps together with different weights to exploit their respective advantages. We can see the overall accuracy improves by a further 0.2\%.  


\begin{table}[t!]
\fontsize{7pt}{8pt}\selectfont\centering
\tabcolsep=1.8pt
\begin{tabular}{|c|ccccccc|c|c|}
	\hline
	& Head & Shoulder & Elbow & Wrist & Hip & Knee & Ankle & Total & AUC \bigstrut\\
	\hline\hline
	FCN\_32s & 93.7 & 85.2 & 74.4 & 65.2 & 86.2 & 81.2 & 77 & 80.4 & 50 \bigstrut\\
	FCN\_16s & 93.9 & 85.9 & 75.1 & 68.3 & 86.3 & 83.4 & 78.5 & 81.6 & 52.1 \bigstrut\\
	FCN\_8s & 94.2 & 86.2 & 75.8 & 68.8 & 86.5 & 83.8 & 78.3 & 82 & 53.7 \bigstrut\\
	FCN$\_$16s~(Extra) & 94.2 & 87.5 & 76.8 & 69.2 & 87.5 & 82.8 & 78.4 & 82.4 & 53.2 \bigstrut\\
	FCN$\_$16s~(Fusion) & \textbf{94.3} & 87.7 & 77.0 & 69.5 & 87.6 & 83.4 & 78.6 & 82.6 & 53.2 \bigstrut\\
	FCN$\_$8s~(Extra) & 94.2 & 87.5  & \textbf{77.2} & 69.6 & \textbf{87.2} & 83.5 & \textbf{79.7}  & 82.6 &  \textbf{54.2}  \bigstrut\\
	FCN$\_$8s~(Fusion) &  \textbf{94.3} & \textbf{87.8} & 77.1 & \textbf{69.8}  & \textbf{87.7} & \textbf{83.7} &\textbf{ 79.7} & \textbf{82.8} & \textbf{54.2} \bigstrut\\
	\hline
\end{tabular}%
\vspace{-1mm}
\caption{Evaluation of multi-scale supervision and multi-scale fusion on top of FCN on the LSP testing set with the OC annotation (@PCK0.2) trained on the LSP training set.}
\label{tab:supervision}%
\vspace{-4mm}
\end{table}%

\section{Conclusion}
In this paper, considering that the distributions of the relative locations of joints are very diverse, we propose a two-stage normalization scheme: human body normalization and limb normalization, making the distributions compact and facilitating the learning of spatial refinement models. To validate the effectiveness of our method, we connect the refinement model to various state-of-the-art joint detectors. Experiment results demonstrate that our method consistently improves the performance on different benchmarks.

{\small
\bibliographystyle{ieee}
\bibliography{egbib}
}

\end{document}